\documentclass[lettersize,journal]{IEEEtran}
\usepackage{amsmath,amsfonts}
\usepackage{algorithmic}
\usepackage{algorithm}
\pdfoutput=1
\usepackage{array}
\usepackage[caption=false,font=normalsize,labelfont=sf,textfont=sf]{subfig}
\usepackage{textcomp}
\usepackage{stfloats}
\usepackage{url}
\usepackage{verbatim}
\usepackage{graphicx}
\usepackage{cite}
\usepackage{color}
\begin{document}

%\title{Slip Detection for Robotic Grasping with Visuo-Tactile Fusion}
\title{Visuo-Tactile-Based Slip Detection Using A Multi-Scale Temporal Convolution Network}

\author{Junli Gao, Zhaoji Huang, Zhaonian Tang, Haitao Song, and Wenyu Liang

\thanks{J. Gao, Z. Huang, and Z. Tang are with the School of Automation, Guangdong University of Technology, Guangzhou 510006, China.}
\thanks{H. Song is with the School of Business Administration, South China University of Technology, Guangzhou 510640, China.}
\thanks{W. Liang is with the Institute for Infocomm Research (I$^2$R), Agency for Science, Technology and Research (A*STAR), 138632, Singapore.} 

        % <-this % stops a space
}

%\markboth{Journal of \LaTeX\ Class Files,~Vol.~, No.~, ~2022}%


\maketitle

\begin{abstract}
Humans can accurately determine whether the object in hand has slipped or not by visual and tactile perception. However, it is still a challenge for robots to detect in-hand object slip through visuo-tactile fusion. To address this issue, a novel visuo-tactile fusion deep neural network is proposed to detect slip, which is a time-dependent continuous action. By using the multi-scale temporal convolution network (MS-TCN) to extract the temporal features of visual and tactile data, the slip can be detected effectively. In this paper, a 7-dregree-of-freedom (7-DoF) robot manipulator equipped with a camera and a tactile sensor is used for data collection on 50 daily objects with different shapes, materials, sizes, and weights. Therefore, a dataset is built, where the grasping data of 40 objects and 10 objects are used for network training and testing, respectively.
%We use the D455 camera, XELA tactile sensor, and XArm 7DoF robotic arm to collect data on 50 daily objects with different shape, material, size, and weight. 
%A public dataset is built. The grasping data of 40 and 10 objects are used for network training and testing, respectively. 
The detection accuracy is 96.96\% based on the proposed model. Also, the proposed model is compared with a visuo-tactile fusion deep neural network (DNN) based on long short-term memory network (LSTM) on the collected dataset and a public dataset using the GelSight tactile sensor. The results demonstrate that the proposed model performs better on both dataset. The proposed model can help robots grasp daily objects reliably. In addition, it can be used in grasping force control, grasping policy generation and dexterous manipulation. %All codes and dataset are available publicly at https://github.com/jonygao621/cnn\_mstcn. 
\end{abstract}


\section{Introduction}
\IEEEPARstart{R}{obotic} grasping stability is critical to executing given tasks, which mostly determines the grasping success rate. When the grasping force is insufficient or the grasping strategy is not suitable, the object will easily slip off and thus lead to grasping failure. Therefore, detecting the robotic grasping state can help the robot adjust the grasping force and strategy to achieve stable grasping \cite{Stachowsky2016,Feng2020}. 

Humans take advantage of multiple modalities naturally when grasping objects, especially in visual and tactile perception. The sense of vision can help humans identify and locate objects quickly. The sense of touch (i.e., tactile sensing) can provide rich contact information to detect sliding or not \cite{Howe1993}. Benefiting from the rapid development of tactile sensors \cite{Yousef2011}, the researches in robotic grasping state detection \cite{Chen2018} have received extensive attention. In \cite{Liu2022}, Liu \textit{et al}. designed a new tactile sensor to improve the accuracy of sliding detection by detecting the contact area. In \cite{Li2018}, Li \textit{et al}. proposed a slip detection method based on LSTM, where an RGB camera and a GelSight tactile sensor are used. The visual and tactile data of the 94 daily objects are collected when the objects are grasped and lifted. The slip detection accuracy is 88.03\%. In \cite{Zapata-Impata2019a}, Zapata \textit{et al}. used the Convolutional LSTM (ConvLSTM) for slip detection, including the types of slippage-translational/rotational and direction. The maximum detection accuracy is 82.56\%. In \cite{Begalinova2020}, Begalinova \textit{et al}. presented a slip detection method based on low-cost tactile sensors which got over 95\% and 89\% accuracy in offline and online classification, respectively. Therefore, it is beneficial to promote the application of tactile sensors in robots for stable object grasping.

Slip is a time-dependent continuous action. It is very important to extract the corresponding temporal features of the visual, tactile and other modal data. In \cite{Zapata-Impata2019}, Zapata \textit{et al}. generated data on 51 objects for training and compare the performance of LSTM and ConvLSTM on slip detection by the BioTac SP tactile sensor. It shows that the LSTM performs better than ConvLSTM. In recent research, recurrent neural networks (RNN), such as LSTM, are used more often for temporal feature extraction. But the fact is that RNN is not easy to train. In 2017, Lea \textit{et al}. proposed a temporal convolutional network (TCN) based on a convolutional neural network (CNN) to extract the temporal features \cite{Lea2017}. The TCN adopts fully convolutional architectures and has faster converging speed with longer temporal memory \cite{Ma2021}. Moreover, Bai \textit{et al}. compared the performance of TCN, LSTM and Gated Recurrent Unit (GRU) comprehensively aimed at the dataset with multiple time-series \cite{Bai2018}. It shows that the TCN outperforms the recurrent network (such as LSTM), across a diverse range of datasets. Up to now, the superior performance of TCN has received extensive attention. Martinez \textit{et al}. presented one MS-TCN based on TCN, which further improves the performance of TCN \cite{Martinez2020}. Ma \textit{et al}. proposed one Densely Connected TCN (DC-TCN) for lip recognition \cite{Ma2021}. It outperforms previous RNNs and achieved state-of-the-art accuracy on the LRW dataset \cite{Chung2017}. 

Considering the excellent performance of MS-TCN, it is employed to extract the temporal features of visual and tactile data from robotic grasping in this work. During the sliding/slip on grasping an object with uncertain speed and displacement in different time periods, the influence of features on different time-dimension may be different. Therefore, the MS-TCN is used to extract the fused multi-dimensional visuo-tactile features after the spatiotemporal feature extraction of input data.

In this paper, a deep neural network for visuo-tactile fusion (CNN-MSTCN) to detect the slip state during robotic grasping is proposed. In recent years, some researchers have shared their dataset for robotic grasping \cite{Li2018, GangYan2022, Wang2019}. However, the acquisition and representation of tactile data depend heavily on the tactile sensor type itself. It is difficult to generalize among different sensors. Here, grasping and lifting experiments using a XELA tactile sensor and a RealSense D455 camera on 50 daily objects using different grasping-force are conducted for collecting tactile and visual data. Significantly, the grasping data of 40 objects are used for training, and the other 10 for testing, respectively. 

This paper aims to deal with the aforementioned issues, and the main contributions are summarized as follows:
\begin{itemize}
    \item  A novel visuo-tactile fusion deep neural network (CNN-MSTCN) is proposed to detect the slip state. The experiments prove that our model has a better performance than the model in \cite{Li2018}.
  
  \item  A multi-modal sensing system is developed and integrated with a robot manipulator for object-stable grasping.
  
 \item  A visuo-tactile dataset is built. It covers rectangular prisms, spheres, cylinders and their composites, a total of 50 representative daily objects with varied sizes, materials and weights in four categories. %It is available at https://github.com/ZhaoJi-Huang/Visuo-Tactile-Based-Slip-Detection-Using-Multi-Scale-Temporal-Convolution-Network.
  
 % \item  Test our model on the dataset from \cite{Li2018}. The results demonstrate that our model has better performance and generalization ability for array-shaped and optical tactile sensors.
\end{itemize}

The rest of this paper is organized as follows. Section \ref{sec:related work} introduces the related works. Section \ref{sec:problem statement and methodology} shows the problem statement and methodology in detail. Where the slip detection problem is formulated first and followed by the presentation of the proposed CNN-MSTCN model based on the methodology of spatiotemporal feature extraction. Section \ref{sec:experiment} analyzes the experimental results before the conclusion in Section \ref{sec:conclusion}.

\section{Related Works}
\label{sec:related work}
\subsection{Slip Detection with Tactile Modal}
In \cite{R.S.JohanssonandA.Vallbo1979}, G. Westling \textit{et al}. implemented the tactile experiments on volunteers, which shows that human tactile feedback is extremely important for grasping. Without tactile feedback, it is difficult for humans to maintain stable grasping. Similarly, the force/tactile feedback information is also important for robotic grasping and slip detection tasks. In  \cite{Jiang2022}, Jiang \textit{et al}. used the GelSight tactile sensor to help robots grasp transparent objects, which is difficult for vision to do. In the past few decades, numerous researchers have conducted extensive research on slip detection using different tactile sensors \cite{Romeo2020}. In \cite{Chen2018a}, Chen \textit{et al}. used a tactile sensor to calculate the minimum grasping force needed to securely grasp an object from the view of friction. However, it has certain limitations when considering the contact surfaces' coefficients of friction. In \cite{Yuan2015}, Yuan \textit{et al}. proposed a method to detect the normal, shear and torsional load on contact surfaces with a GelSight tactile sensor. It is proved that the slip can be detected on a human-like robot gripper. In \cite{Dong2019}, Dong \textit{et al}. used the GelSlim tactile sensor to detect the sensor displacement field during grasping, which is used to detect the initial slip. It can achieve good slip-detection results when dealing with rigid objects, but it is not ideal for deformed objects.

To improve the accuracy of slip detection, researchers have used various methods to process tactile data. In \cite{James2020}, James \textit{et al}. used support vector machines (SVM) to detect slip phenomena. This method can deploy on robot arms to respond to slips in real-time while the robots are grasping different objects. In \cite{Yan2021}, Yan \textit{et al}. focused on the temporal features of tactile data. They use CNN with spatial-channel and temporal attention mechanisms to predict grasping stability. This is the first time used an attention mechanism for predicting grasp stability that only relies on tactile information. Furthermore, in \cite{Zapata-Impata2018}, Zapata \textit{et al.} used CNN to process non-matrix tactile sensor information for slip detection. To take full advantage of the location information of non-matrix tactile sensors, Garcia \textit{et al}. used graph convolutional network (GCN) to process tactile information in \cite{Garcia-Garcia2019}.

Most of the above studies are based on tactile data for slip detection. But for objects with smooth surfaces, it becomes less reliable. The human can recognize it when the slippage occurs by means of vision and touch. In addition to the geometric features of objects, vision can also intuitively provide slip information, such as the state and direction of the slip. Therefore, it is necessary to pay attention to visuo-tactile information and make full use of both modal features to improve the accuracy of slip detection.

\subsection{Slip Detection with Visuo-Tactile Fusion}
Due to the differences in the structure, form, and characteristics of visual and tactile data, it is still challenging to detect grasping states by the two modalities. In \cite{Calandra2017} Calandra \textit{et al}. investigated whether tactile sensors can help to predict grasping stability with a multi-modal vision and touch sensing framework. The experimental result shows that the incorporated tactile readings greatly improve grasping performance. In \cite{Jara2014}, Jara \textit{et al}. proposed a robotic system, integrating visual and tactile information and using a model-based approach, to execute grasping tasks, which effectively improves grasping performance by using vision-only or touch-only modal input. In \cite{Cui2020}, Cui \textit{et al}. presented a novel 3D CNN to achieve spatiotemporal feature extraction from visual and tactile data. For the visuo-tactile dataset collected with a XELA tactile sensor and an RGB camera, a prediction accuracy of 99.97\% is achieved. Then, in \cite{Cui2020a}, Cui \textit{et al}.  used the transformer structure to fuse visual and tactile modalities. The result shows that their proposed model can achieve a prediction accuracy of 97.43\%. It is also effective on the dataset in \cite{Li2018} with better generalization performance. However, no matter whether using 3D CNN to extract spatiotemporal features in \cite{Cui2020} or transformer structures to fuse visual-touch features in \cite{Cui2020a}, a large number of model parameters are inevitable, which makes the model difficult to be trained. In addition, to deploy on a real robot, too many model parameters will lead to higher hardware requirements, which is not conducive to different working environments. As for our proposed model, the MS-TCN is used to extract temporal features and fuse visuo-tactile features. There are only 0.08M parameters in one MS-TCN module, which will facilitate in training and deploying the model.

In this paper, the visual and tactile data for slip detection are used. The pre-trained network Resnet34 is used to extract the spatial features of visual data and a three-layer CNN is used to extract spatial features of tactile data. In addition, the MS-TCN is used to extract the temporal features of visual and tactile data. Furthermore, the MS-TCN is also used to fuse the spatiotemporal features of visual and tactile data. Finally, it predicts whether there is a slip or not through a fully connected layer.

\section{Problem Statement and Methodology}
\label{sec:problem statement and methodology}

\subsection{Problem Statement}
The goal of this work is to obtain the grasping state by visuo-tactile fusion information when a robot manipulator grasps the daily objects. The visual and tactile input sequences are set as $x_v^T$, and $x_t^T$, respectively. Then $x_v^T$, $x_t^T$ are fed into the visual and tactile feature extraction model $F_v$, $F_t$, to get their spatiotemporal features $T_v$, $T_t$, respectively. The fused features $T_{(v,t)}$ after concatenation are fed into the network $F_{(v,t)}$ to predict the grasp state $y$. This problem is formulated as follows:

\begin{equation}
x_v^T=(x_v^0,x_v^1,...,x_v^{(T-1)},x_v^T )
\label{eq:01}
\end{equation}
\begin{equation}
x_t^T=(x_t^0,x_t^1,...,x_t^{(T-1)},x_t^T) 
\label{eq:02}
\end{equation}
\begin{equation}
T_{(v,t)}=F_v{(x_v^T)} \oplus F_t {(x_t^T)}
\label{eq:03}
\end{equation}
\begin{equation}
y=F_{v,t}{(T_{(v,t)})}, y \in (0,1)
\label{eq:04}
\end{equation}

Here, $0$ and $1$ refer to the slip and stable state, correspondingly. It means the assessment of the grasping state is defined as a binary classification problem. $T$ represents the length of input sequences, where $x_v^T$ and $x_t^T$ have same length.

\begin{figure*}[!t]
\centering
\subfloat[TCN]{\includegraphics[width=2.5in]{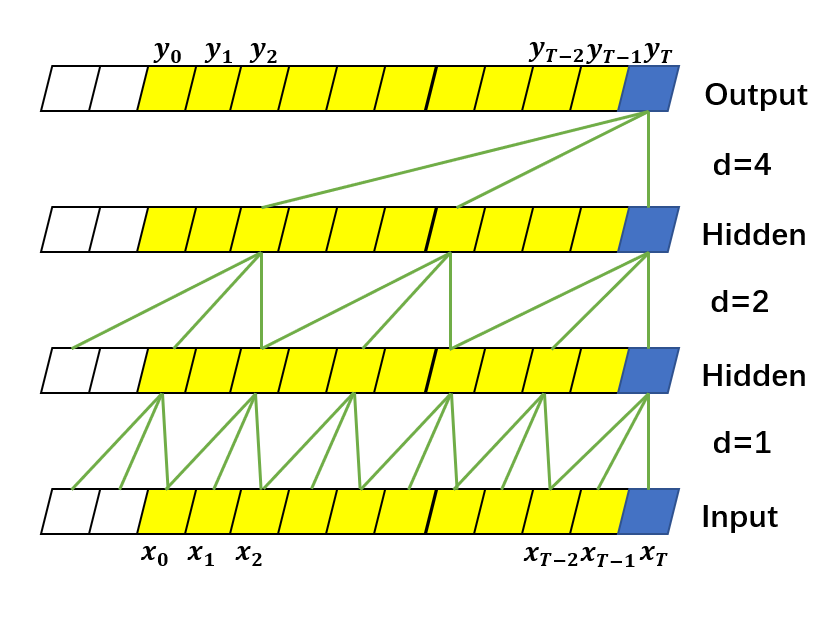}%
\label{fig:TCN model}}
\hfil
\subfloat[MS-TCN]{\includegraphics[width=2.5in]{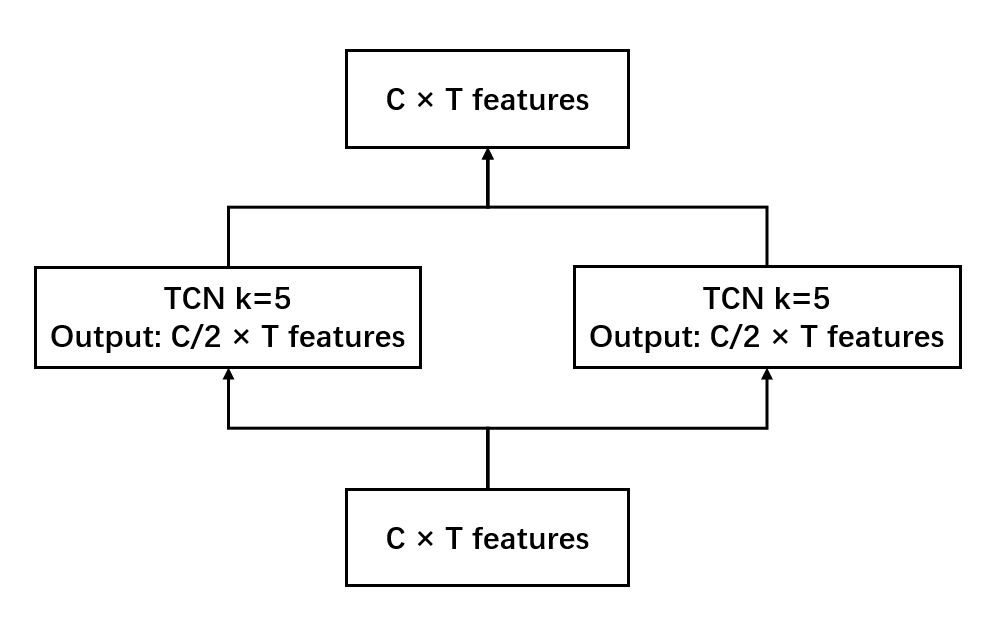}%
\label{fig:MS-TCN model}}
\caption{TCN and MS-TCN: (a) The diagram of TCN \cite{Bai2018}. The dilated causal convolution with dilation factors $d$ is 1, 2, 4 and filter size $k$ is 3. The receptive field can cover all values from the input sequences. (b) The diagram of MS-TCN.}
% \label{fig_sim}
\end{figure*}

\subsection{Spatial Feature Extraction}
Resnet can effectively solve the degradation problem of a deep network, that is, the deeper the network layer, the worse performance. The residual structure enables Resnet to stack deeper layers, such as 18, 34, 50, and 101, to have excellent performance. Resnet34 \cite{He2016} has a stronger ability on feature extraction after training on ImageNet \cite{Deng2010} and it is widely used in transfer learning. In addition, the other common pre-trained network trained on ImageNet is Resnet50 \cite{He2016}, VGG-16 \cite{Simonyan2015}, Inception-V3 \cite{Szegedy2016}, and so on. The feature extraction capabilities of the pre-trained networks will vary on different datasets. Through experimental comparisons, it is found that Resnet34 performs better on our dataset. Therefore, this paper uses Resnet34 to extract the spatial features of visual image sequences. But Resnet34 is not suitable for tactile image sequences whose size is only $4 \times 4$. Therefore, one three-layer CNN is built to extract the spatial features with 32-dimension, where the specific network parameters are shown in Table \ref{tab: parameters of tactile feature extraction network}.

\begin{table}[!t]
	\caption{The parameters of tactile feature extraction network} \label{tab: parameters of tactile feature extraction network}
	\centering
	\begin{tabular}{|c | c | c |}
% 		\toprule
		\hline
		\textbf{Layer}	& \textbf{Tactile input $(4\times4\times3)$}	& \textbf{Output size} \\
		\hline
		con1  & $3\times3\times8$, padding(1,1), stride(1,1), relu & $4\times4\times8$ \\
		pool1 & max(2,2), stride(2,2) & $2\times2\times8$ \\
		\hline
		con2  & $3\times3\times16$, padding(1,1), stride(1,1), relu & $2\times2\times16$ \\
	    pool2 & max(2,2), stride(2,2) & $1\times1\times16$ \\
	    \hline
		con3  & $1\times1\times32$, padding(1,1), stride(1,1), relu & $1\times1\times32$ \\
	    pool3 & max(1,1), stride(1,1) & $1\times1\times32$ \\
	    \hline
	\end{tabular}
\end{table}

\subsection{Temporal Feature Extraction}
The structure of TCN is shown in Fig. \ref{fig:TCN model}, including four layers. The size of each convolution kernel is 3, and $d$ represents the dilation factor. Due to the dilated convolutions, only four layers are needed to allow the receptive field of the top output to cover all the input ranges. Each layer of TCN is a 1D convolutional network, and the features of all time points can be processed in parallel through it. Here, 1D convolution can look back at the history with the kernel size. It is essential for a deeper network to sequence tasks, especially for long history. To address this issue, TCN takes dilated convolutions to expand the receptive field. The MS-TCN is proposed based on TCN to enable the network with visibility into multiple temporal scales, where the short and long-term information can be considered during feature encoding \cite{Martinez2020}. The structure of the MS-TCN is shown in Fig. \ref{fig:MS-TCN model}. Each layer has $n$ branches, and each branch has $C/n$ convolution kernels. $C$ represents the channel dimensionality of the layer. The output of each layer is obtained by concatenating the outputs of all branches. In this way, every convolution layer mixes the information at several temporal scales. In the proposed model, the MS-TCN is employed to extract the temporal features of visual and tactile data.

\section{Model Description}
\begin{figure}[!t]
	\centering
	\includegraphics[width=8cm]{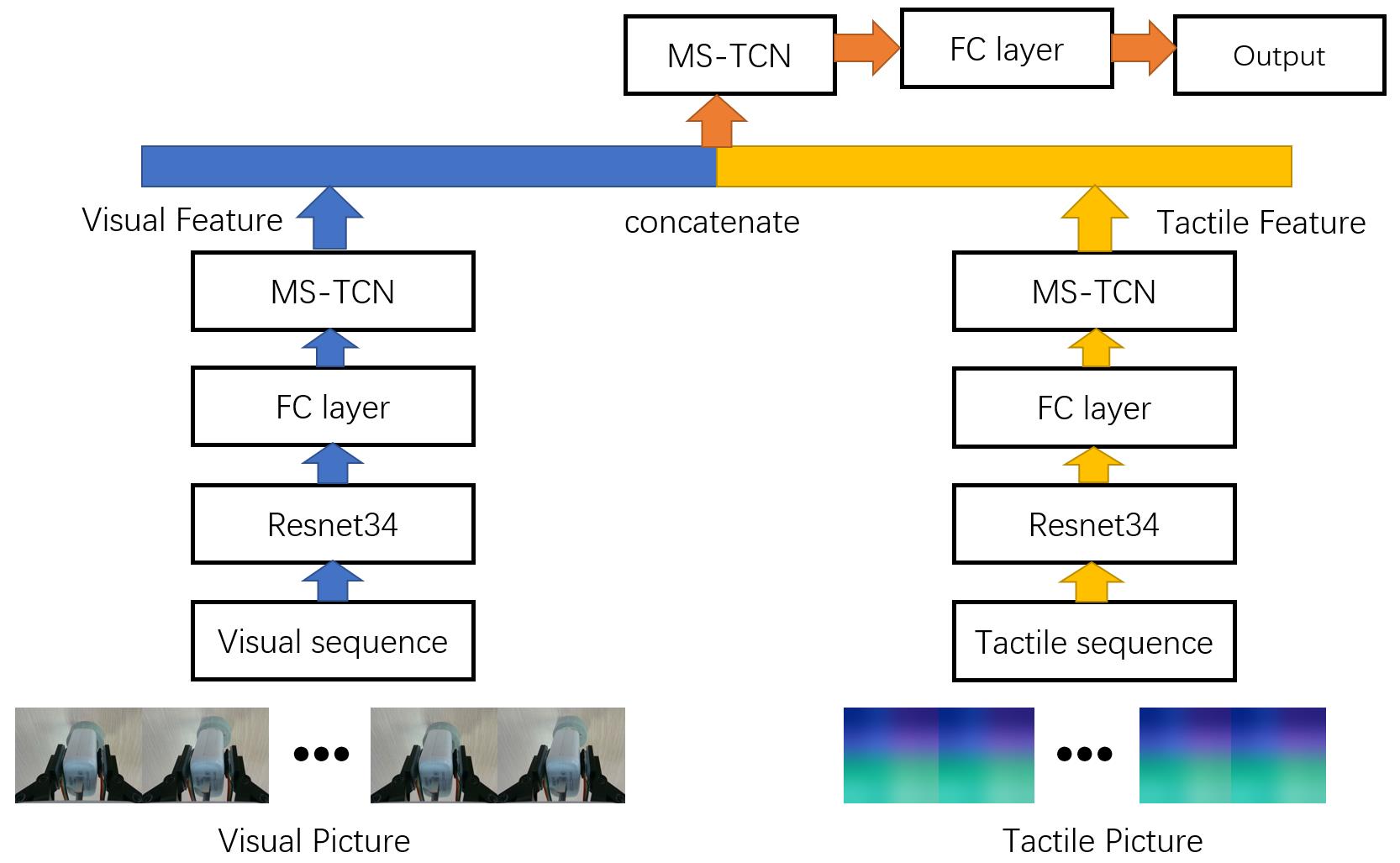} 
	\caption{The structure of our CNN-MSTCN model. Left: The visual input is fed into Resnet34 and MS-TCN for spatiotemporal feature extraction. Right: The tactile input is fed into tactile feature extraction and MS-TCN to extract the spatiotemporal features. The visual and tactile spatiotemporal features are concatenated, and then fed into MS-TCN. The output of the fusion features through MS-TCN are fed into FC layer for binary classification.}
	\label{fig:CNN-MSTCN model}
\end{figure}

To address the aforementioned problems, a CNN-MSTCN model, including visuo-tactile feature extraction and fusion modules, is presented and shown in Fig. \ref{fig:CNN-MSTCN model}. The input to the model consists of two modalities, namely, visual picture and tactile picture. The length of both input sequences is 13. The output is fed into a fully connected (FC) layer to get the binary classification result, where $0$ and $1$ represent the slip and stable state, respectively.

The spatial features of the visual image sequence are extracted by the pre-trained Resnet34. Considering the image size in ImageNet is $224 \times 224$, it needs to be resized to the same size. The obtained spatial features are fed into an FC layer to reduce dimensionality to 64, and then the temporal features are extracted by the MS-TCN. For the tactile image sequence, it is fed into the feature extraction module to obtain 32-dimensional tactile features, which are upgraded to 64 through an FC layer. Then, the 64-dimensional tactile features are fed into the MS-TCN to extract the temporal features. For the visual and tactile feature extraction, the structure of the MS-TCN is the same. There are two layers, and each layer has two branches. The size of the convolution kernel of each branch is 5, and the output feature dimensionality of each layer is 64. After feature extraction, we can get the $b \times 64 \times n$ dimensional features from the visual and tactile image sequences, respectively, where $b$ is the batch size and $n$ is the image sequence length. The visual and tactile features are concatenated to obtain $b \times 128 \times n$ dimensional features. The fused features after concatenation are fed into the MS-TCN to extract the relationship among features in different time dimensions. Here, the MS-TCN has three layers, and each layer has three branches. The size of the convolution kernel of each branch is 3, and the feature dimension of each layer is 64. Finally, the output of the MS-TCN is fed into an FC layer for classification.

Considering our dataset that is relatively small, the pre-trained Resnet34 is frozen and only the rest of the model is trained. The model is built based on the Pytorch platform, and network parameters are randomly initialized. The cross-entropy function is used as a loss function. The learning rate of the Adam optimizer is $1\times 10^{-7}$, and the batch size is 8. The host computer is equipped with an Intel Xeon E5-2620 CPU, 48GB memory, and two GTX1080Ti GPUs.

\section{Experiments and Results}
\label{sec:experiment}

\subsection{Experimental System Setup}
\begin{figure}[htbp]
    \centering
    \includegraphics[width=8cm]{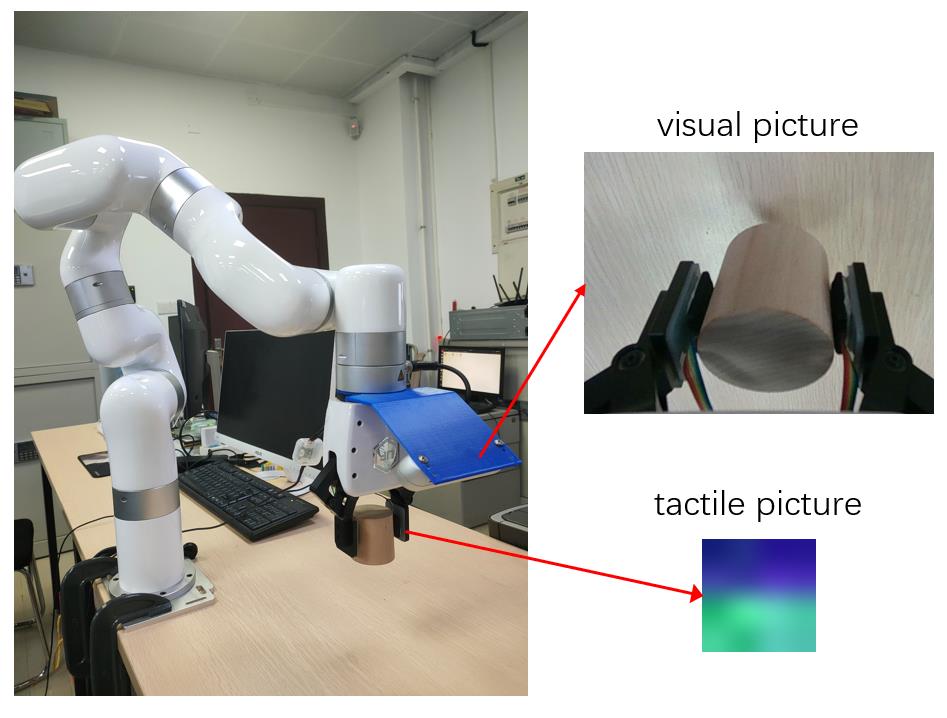} 
    \caption{Left: The experimental system setup: xArm 7DOF robotic arm with 2-finger gripper. Each finger of the gripper is equipped with a XELA tactile sensor. The D455 camera is mounted on the top-side of the gripper. Upper right: visual picture captured by the wrist camera. Bottom right: tactile picture converted from tactile readings.}
    \label{fig:data acquisition diagram}
\end{figure}

In this work, a xArm 7-DoF robotic arm with a 2-finger gripper, a RealSense D455 RGB camera, and two XELA tactile sensors are used to build the data acquisition platform, which is shown in Fig. \ref{fig:data acquisition diagram}. The XELA tactile sensors are attached to the surface of the silicone pads of the gripper's fingers, and the D455 camera is mounted on the upside of the robot gripper in the form of ``eye-in-hand". The maximum displacement of the gripper's fingers changes from 86 mm to 77 mm due to the tactile sensors' setup. The original size of the RGB pictures captured by the camera is $1280 \times 720$ pixels, which are tailored to $640 \times 480$ as the visual images. One XELA sensor has 16 force sensing points, i.e., $4 \times 4$. Each sensing point measures 3D force-tactile data along the X/Y/Z-axis. The tactile data are all converted into RGB images of $4 \times 4$ pixels. All the devices work on the ROS Melodic.

\subsection{Data Acquisition}
\begin{figure*}[!t]
	\centering
	\includegraphics[width=15cm]{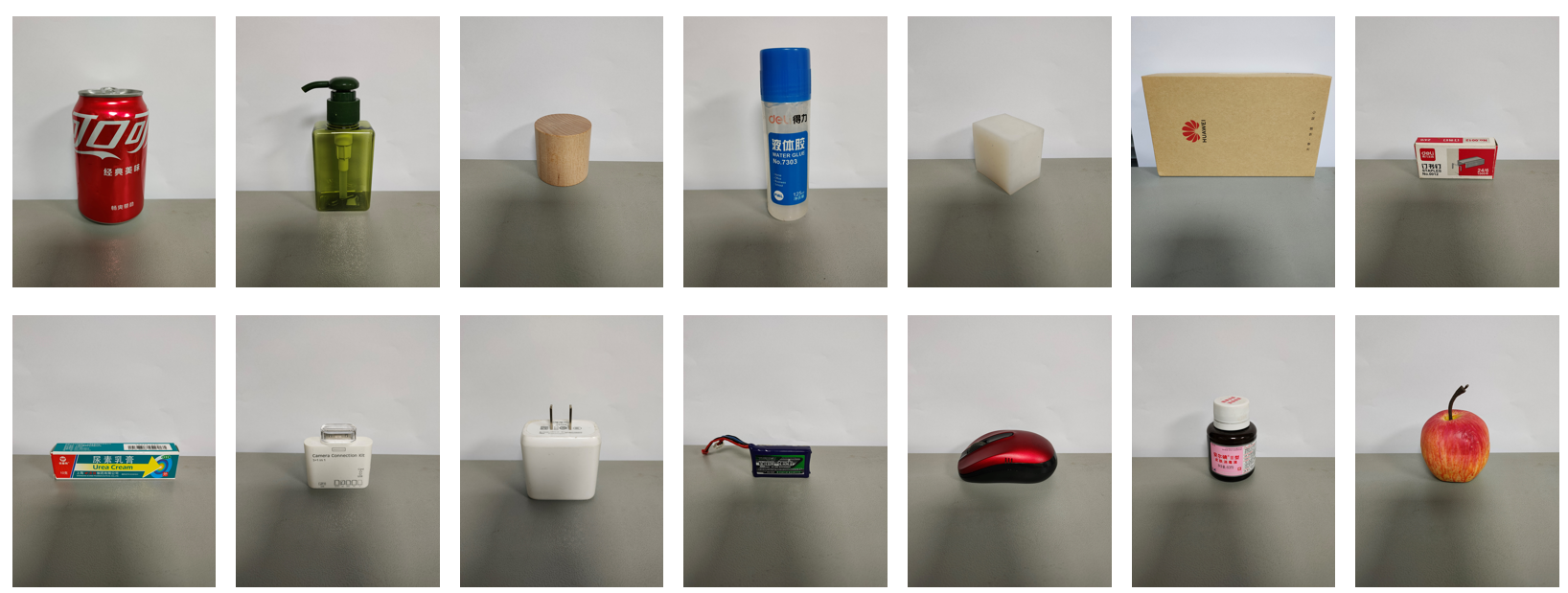}
	\caption{Part of samples in our dataset. The selected objects have different sizes, shapes, materials and weights.} 
    \label{fig:some grasp objects}
\end{figure*}

50 representative daily objects in four categories of rectangular prisms, spheres, cylinders and their complexes are selected for data acquisition on the robotic grasping task. Some of the selected objects are shown in Fig. \ref{fig:some grasp objects}. The 50 objects are varied in size, shape, material, and weight. All of their widths are less than the maximum displacement of the gripper's fingers, i.e., 77 mm so that all the objects can be grasped by the gripper. Considering the measuring range of the XELA tactile sensors, all of their weights are under 1 kg. Inspired by the literature \cite{Li2018}, it is necessary to determine the critical grasp width of each object at first, which indicates that the object can just be grasped stably. The practical grasp width is presupposed to balance the stable and slidable grasp labels based on the critical value.

During the data collection, each object needs to be placed in a predetermined position in advance. Firstly, the robot manipulator is moved to the grasping point according to the preset grasping position and gripper's opening width. Secondly, the data acquisition is started and the gripper is closed at the same time. Then, the robot manipulator raises up the gripper by 30 mm and then the data acquisition is stopped. Finally, the robot manipulator moves down the gripper by 30 mm and returns to the grasping point to release the grasped object by the gripper. Each object will be grasped and lifted 20 times. Each grasp width is preset and the tag file is generated automatically. The dataset includes the visual and tactile data captured by the camera and the tactile sensor, respectively. These two modal data are corresponding in time, that is, one RGB image and one tactile data are collected at the same time. The sampling frequency is 30 Hz, so about 40 frames of data can be obtained in one round of data collection. Considering the response time on the start and stop of the robot manipulator, the data can be still collected during this period, and the final valid data obtained by one grasp are 20 frames. At last, our dataset contains 952 raw valid grasp data. On this basis, referring to the data augmentation method introduced by Li \textit{et al}. in \cite{Li2018}, the data are augmented by 9 times. Hence, a total of 8568 groups of image sequences are obtained, including 13 frames per group. Therein, the 6795 sets of data on 40 objects are used as a training set, and the other 1773 sets of 10 objects are used as a testing set. Source code and data set is available at https://github.com/ZhaoJi-Huang/Visuo-Tactile-Based-Slip-Detection-Using-Multi-Scale-Temporal-Convolution-Network.

\subsection{Experimental Results}
To evaluate the performance of the CNN-MSTCN proposed in this paper, our dataset is used for training and testing. The main factors that affect the performance include the pre-trained model, length of input sequences, input modal, and so on. The details are discussed as follows.

\subsubsection{Pre-trained Model} To compare the feature extraction ability of different pre-trained models on visual data, Resnet18, Resnet34, Resnet50, VGG-16 and Inception-V3 models are used for testing successively. All the pre-trained weight parameters for these models are obtained based on ImageNet. The visual images are resized to $224 \times 224$ and the length of the input sequence is preset to 13. The input modalities are visuo-tactile fusion. The test results are shown in Table \ref{tab:results of network pre-training}. The different pre-trained models do have an impact on the feature extraction effect of visual data, which may be related to the distribution of data itself. Finally, Resnet34 is selected to extract the spatial features of visual data in the follow-up tests as it performs best among all the models.

\begin{table*}[!t]
	\caption{The test result of different pre-trained network} \label{tab:results of network pre-training}
	\centering
	\begin{tabular}{|c | c | c | c | c | c |}
		\hline
		\textbf{Pre-training model}	& \textbf{Resnet18}	& \textbf{Resnet34} & \textbf{Resnet50} & \textbf{VGG-16} & \textbf{Inception-V3} \\
		\hline
		\textbf{Precision} & 92.88\% & \textbf{95.19\%} & 91.65\% & 89.25\% & 86.43\% \\
		\hline
		\textbf{Recall} & 98.56\% & \textbf{99.00\%} & 96.33\% & 98.67\% & 96.22\% \\
		\hline
		\textbf{F1 score} & 95.63\% & \textbf{97.06\%} & 93.93\% & 93.72\% & 91.06\% \\
		\hline
		\textbf{Accuracy} & 95.44\% & \textbf{96.96\%} & 93.69\% & 93.30\% & 90.43\% \\
		\hline
	\end{tabular}
\end{table*}

\subsubsection{Input Sequence Length} 
\begin{figure}[htbp]
	\centering
	\includegraphics[width=8cm]{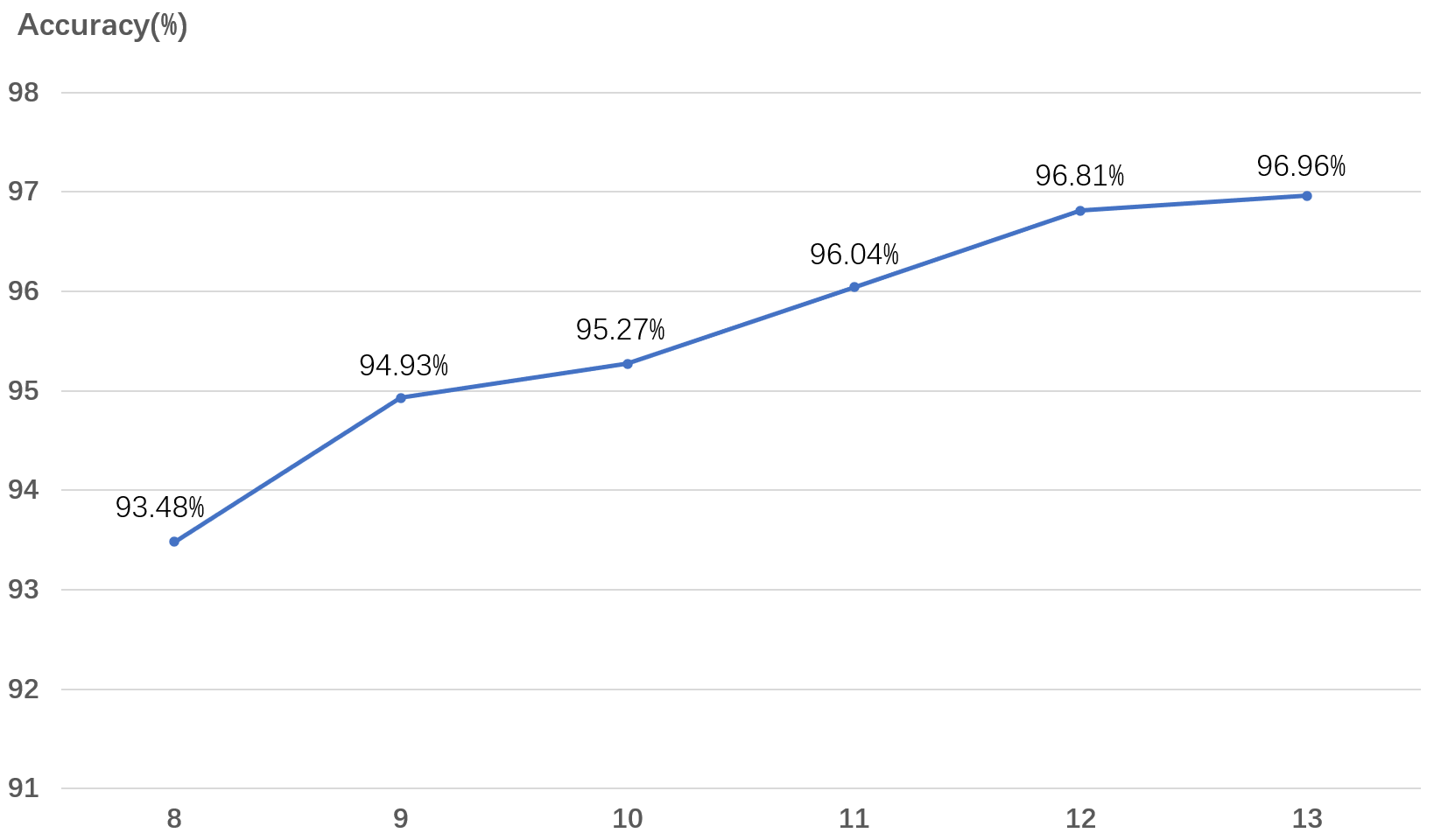} 
	\caption{The accuracy curve of input sequence with different length} 
	\label{fig:the result of different length}
\end{figure}

A longer input sequence means that more temporal information can be fed into the network, but it may also bring more useless feature information. Here, the Resnet34 and visuo-tactile fusion modalities are used. The input sequence length is set to 8, 9, 10, 11, 12, and 13, respectively. It can be seen that longer lengths can get higher accuracy as shown in Fig. \ref{fig:the result of different length}. It means that useful information provided by the longer sequence affects the network more than useless information. For example, when the input sequence length is 13, the accuracy is 96.96\%, which is improved by 3.48\% compared with 8. It can be also observed that the trends of the improvements become stable after the input sequence length of 12. Thus, the input sequence length of 13 will be used in the subsequent experiments.

\subsubsection{Input Modality} The visual data include rich geometric features of objects while the tactile data have detailed contact information. For different objects, the effects of visual and tactile information on the grasping state will also be various. Therefore, the visuo-tactile fusion except for visual-only and tactile-only modalities is considered. For the visual-only input, the Resnet34 and the MS-TCN are used to extract the spatiotemporal features. For the tactile-only, one CNN-based tactile feature extraction and the MS-TCN are used. The classification is achieved through one FC layer for both visual-only and tactile-only input. For the visuo-tactile fusion modality, the CNN-MSTCN is used for feature extraction and classification. The length of the input sequence is all set to 13. The test results are shown in Table \ref{tab:accuracy on different modal}. The performance of visual-only is better than tactile-only. It means that the visual modality plays a more critical role in the object-grasping task. Moreover, the performance of the visuo-tactile fusion is better than visual-only. It implies the tactile modality is also beneficial, and can effectively improve the performance of the overall network.

\begin{table}[htbp]
	\caption{The accuracy on different modal} \label{tab:accuracy on different modal}
	\centering
	\begin{tabular}{|c | c | c | c |}
		\hline
		\textbf{Modal}	& \textbf{Tactile}	& \textbf{Visual} & \textbf{Visual-tactile} \\
		\hline
		\textbf{Precision} & \textbf{100\%} & 92.69\% & 95.19\%  \\
		\hline
		\textbf{Recall} & 51.67\% & \textbf{100\%}\% & 99.00\%  \\
		\hline
		\textbf{F1 score} & 68.13\% & 96.21\% & \textbf{97.06\%}  \\
		\hline
		\textbf{Accuracy} & 75.51\% & 96.00\% & \textbf{96.96\%}  \\
		\hline
	\end{tabular}
\end{table}

\begin{figure*}[!t]
	\centering
	\includegraphics[width=15cm]{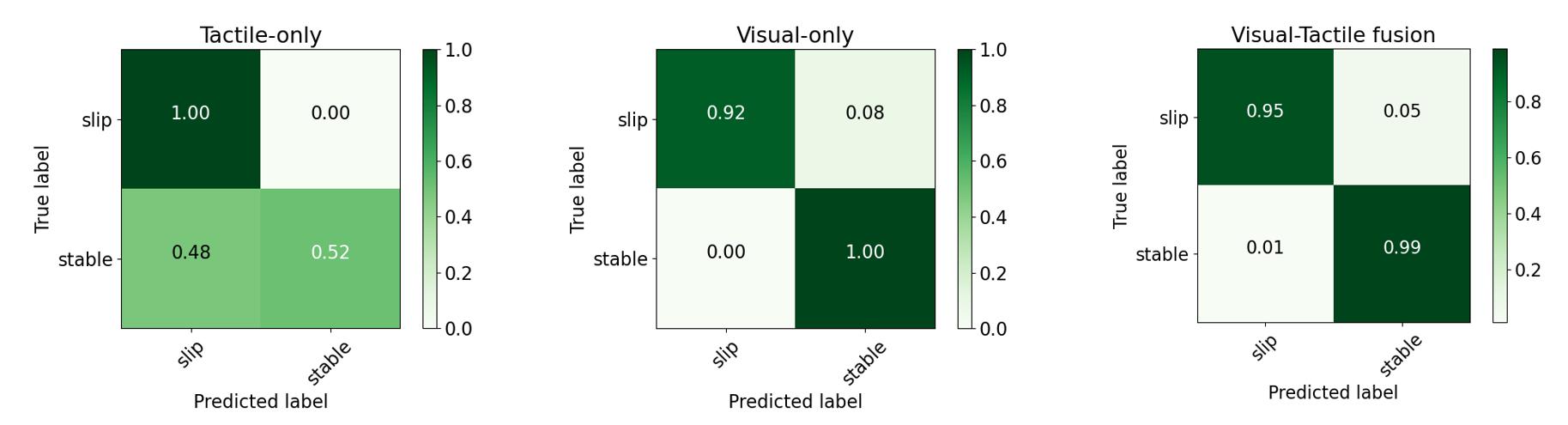} 
	\caption{The confusion matrix of different input modalities. Single-tactile confusion matrix (left), Single-visual confusion matrix (centre), Visual-tactile fusion confusion matrix (right).} 
	\label{fig:different model confusion matrix}
\end{figure*}

The confusion matrices of different input modalities are shown in Fig. \ref{fig:different model confusion matrix}. The tactile-only is accurate for slip detection while the visual-only is very accurate for stability prediction, respectively. However, the tactile modality is not very strong in stability prediction while the visual modality can make up for this. Therefore, visuo-tactile fusion can significantly improve the prediction accuracy of stability. Although the detection ability of visual modality on the slip is not too weak, the addition of tactile modality can further improve the detection accuracy. Therefore, the visuo-tactile fusion can take full advantage of both modalities and improve the overall model prediction and detection abilities effectively.

\subsubsection{Comparison} Through the above three tests, the best strategy profile includes the pre-trained Resnet34, the input sequence length of 13, and visuo-tactile fusion. To further verify the performance of the proposed model, a comparison experiment is conducted to compare the performance of the TCN and the MS-TCN. As can be seen from the results in Table \ref{tab:comparison test with our dataset}, the performance of the MS-TCN is indeed better than that of the TCN. Compared with TCN, the MS-TCN can integrate more dimensional features. In addition, it is compared with the method proposed by Li \textit{et al}. in \cite{Li2018}. To ensure the consistent experimental condition, the pre-trained network used in \cite{Li2018} is changed to Resnet34. Both networks take the Resnet34 and tactile feature extraction model to extract the spatial features of visual and tactile modalities, respectively. Furthermore, the MS-TCN is used to extract the temporal features and improve the visuo-tactile fusion method. The results are shown in Table \ref{tab:comparison test with our dataset}. The accuracy of the proposed model (that is 96.96\%) improves by 3.27\% compared with the model proposed in \cite{Li2018} where the LSTM is used to extract temporal features and realize modal fusion.

\begin{table}[htbp]
	\caption{The comparison test on different network with our dataset} \label{tab:comparison test with our dataset}
	\centering
	\begin{tabular}{|c | c | c | c |}
		\hline
		\textbf{Model}	& \textbf{Network in \cite{Li2018}} & \textbf{CNN-TCN} & \textbf{CNN-MSTCN} \\
		\hline
		\textbf{Precision} & 90.12\% & 93.22\% & \textbf{95.19\%} \\
		\hline
		\textbf{Recall} & 98.33\% & \textbf{99.22\%} & 99.00\% \\
		\hline
		\textbf{F1 score} & 94.05\% & 96.12\% & \textbf{97.06\%} \\
		\hline
		\textbf{Accuracy} & 93.69\% & 95.95\% & \textbf{96.96\%} \\
		\hline
	\end{tabular}
\end{table}

\subsubsection{Generalization}
Finally, to verify the generalization ability of the proposed model, the dataset in \cite{Li2018} is also used. This dataset uses an optical-based GelSight tactile sensor to capture the deformation state when it contacts the surface of an object. The size of the tactile image is $640 \times 480$. As our tactile feature extraction network is designed for $4 \times 4$ images, it is not suitable for GelSight tactile images. Therefore, the Resnet34 is used to extract tactile modality features. Refer to \cite{Li2018}, the length of the input sequence is set to 8. The experimental results are shown in Table \ref{tab:comparison test with the open dataset}. Although the tactile sensor used in this dataset is different from ours, the performance of our proposed model still performs better than the model proposed in \cite{Li2018}. It can be concluded that our model can perform well on either dataset using array-shaped or optical tactile sensors. This can help to deploy the proposed model on different robotic systems where different types of tactile sensors are equipped.

\begin{table}[htbp]
	\caption{The comparison test on different network using the dataset in \cite{Li2018}} \label{tab:comparison test with the open dataset}
	\centering
	\begin{tabular}{|c | c | c |}
		\hline
		\textbf{Model}	& \textbf{CNN-MSTCN}	& \textbf{Network in \cite{Li2018}} \\
		\hline
		\textbf{Precision} & \textbf{77.54\%} & 76.38\% \\
		\hline
		\textbf{Recall} & \textbf{82.86\%} & 67.43\% \\
		\hline
		\textbf{F1 score} & \textbf{80.11\%} & 71.62\% \\
		\hline
		\textbf{Accuracy} & \textbf{79.28\%} & 73.09\% \\
		\hline
	\end{tabular}
\end{table}

\subsection{Experimental analysis}
\begin{figure*}[!t]
\centering
\subfloat[cola bottle]{\includegraphics[height=5cm,keepaspectratio]{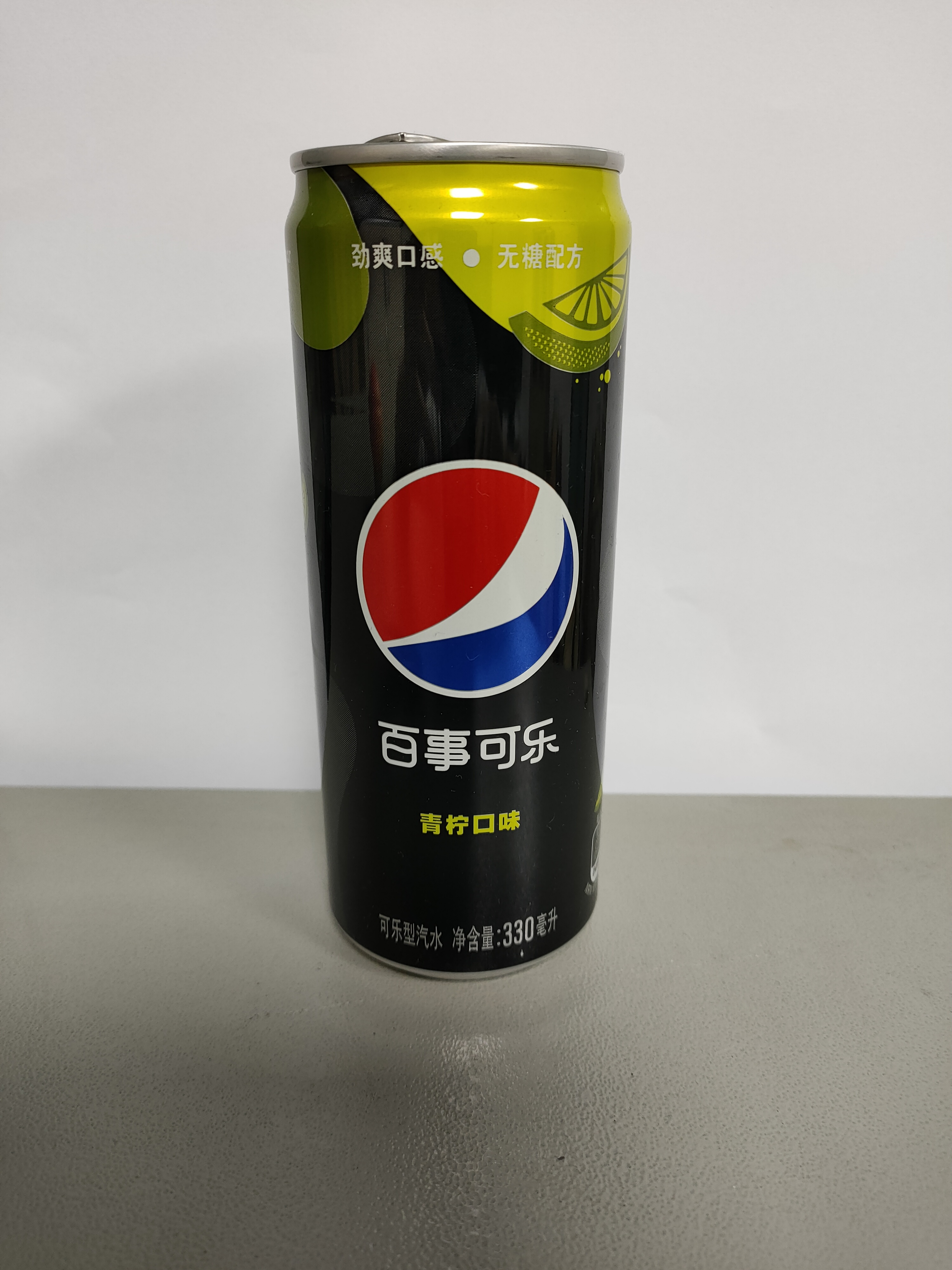}%
\label{fig:cola bottle}}
\hfil
\subfloat[tennis ball]{\includegraphics[height=5cm,keepaspectratio]{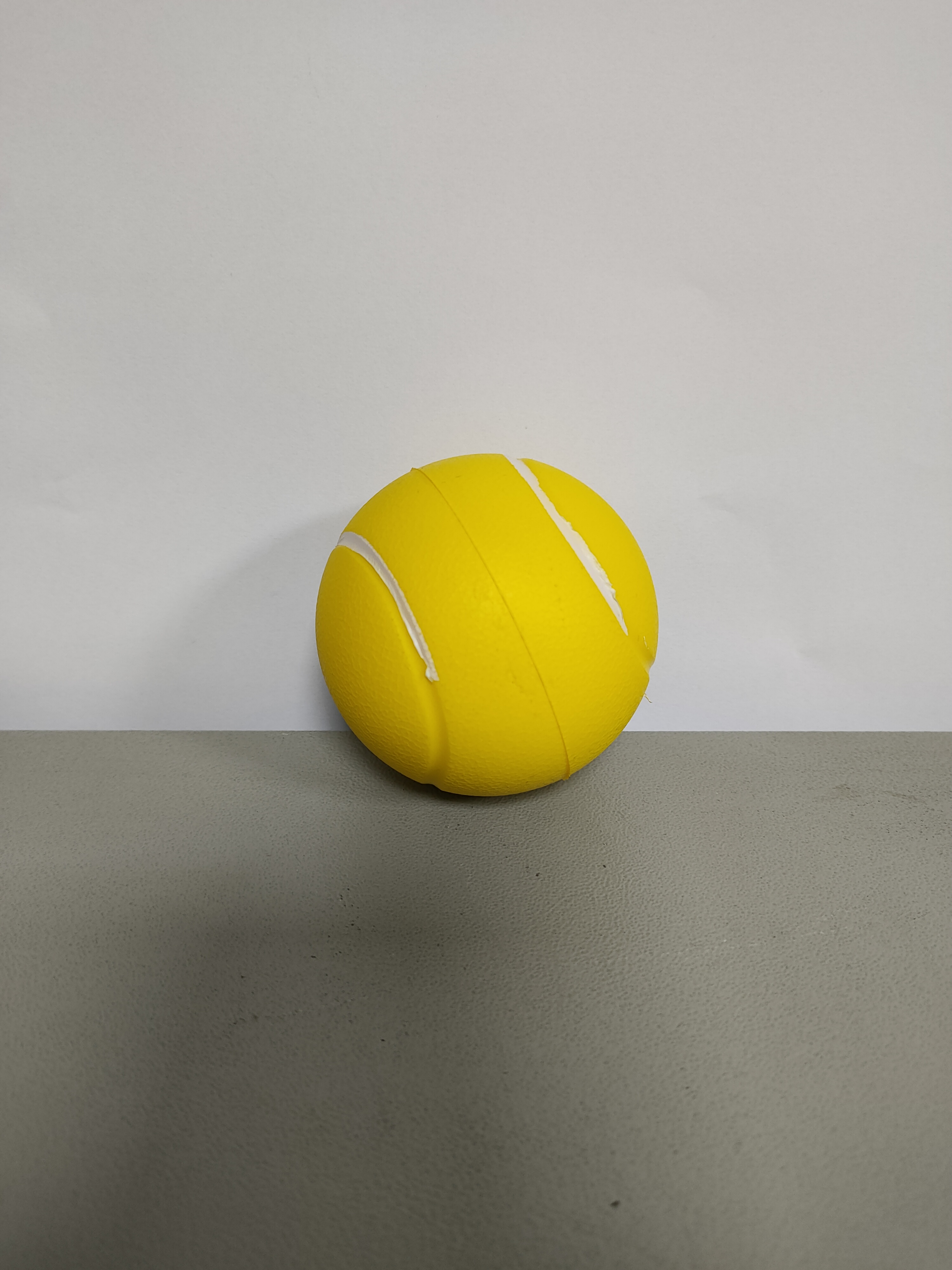}%
\label{fig:tennis}}
\hfil
\subfloat[soft massage ball]{\includegraphics[height=5cm,keepaspectratio]{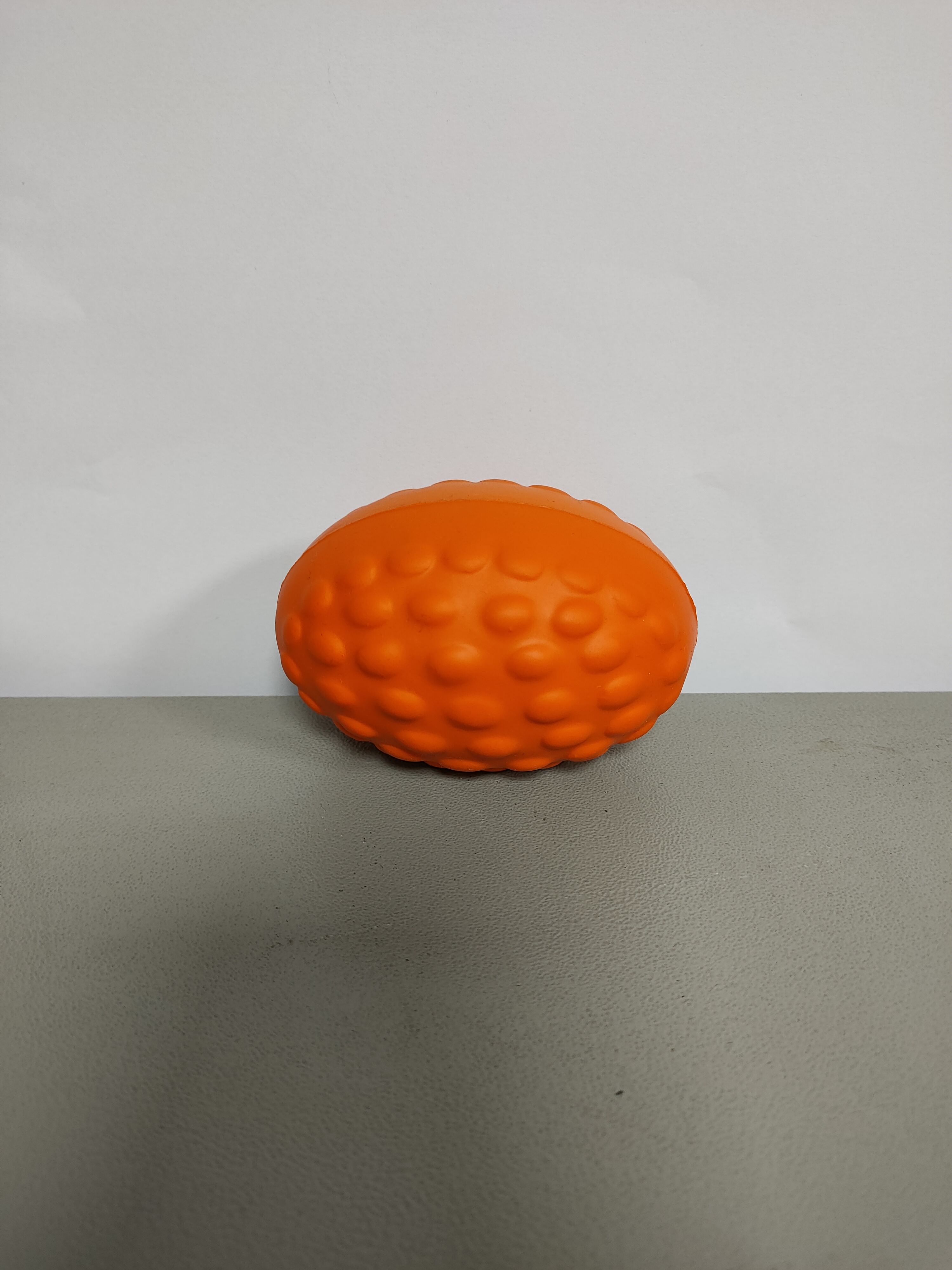}%
\label{fig:soft massage ball}}

\caption{Some objects in the test set: (a) Cola bottle with maximum stiffness; (b) Tennis ball with stiffness between cola bottle and soft massage ball; (c) Soft massage ball with the smallest stiffness.}
\label{fig:some object of test dataset}
\end{figure*}

From the above experimental results, it can be concluded that the visuo-tactile fusion can take full advantage of the two modality information to improve the overall model prediction accuracy. Compared with the visual-only modality, the performance of the tactile-only is not good enough. In this subsection, the reasons are explored. To this end, three objects with different shapes, stiffness and weight in the test set are selected, namely, cola bottle, tennis ball, and soft massage ball, as shown in Fig. \ref{fig:some object of test dataset}.

\begin{figure*}[!t]
	\centering
	\includegraphics[width=15cm]{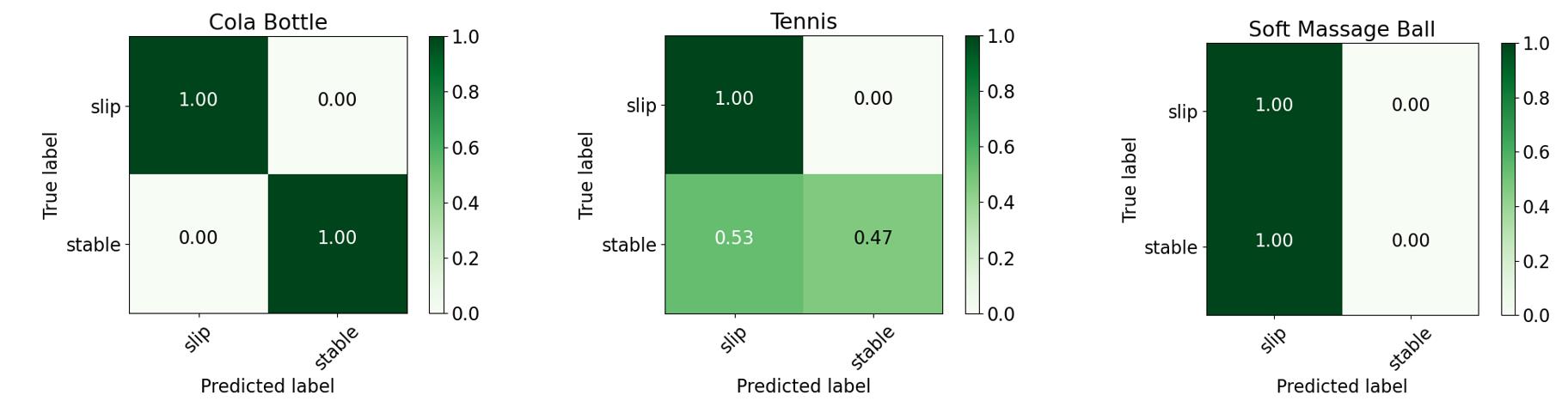} 
	\caption{The confusion matrices: cola bottle (left), tennis (centre), soft massage ball (right).} 
	\label{fig:test confusion matrix}
\end{figure*}

The three objects are tested individually based on the tactile-only input, and the confusion matrices are shown in Fig. \ref{fig:test confusion matrix}. For the cola bottle, the performance is very good with 100\% prediction accuracy. For the tennis ball, the detection ability on the slip is still well, but 53\% of stable grasping states are mistaken as slip. For the soft massage ball, it is disappointing in grasping stability prediction, which takes all the stable states as slip. Obviously, as for different objects, the performance of tactile-only prediction varies greatly. For the cola bottle, the prediction accuracy is 100\%, but only 50\% for the soft massage ball. It is suspected that there may have something to do with the stiffness of the objects. Among the selected objects, the stiffness of the cola bottle is the largest, the tennis ball is the second, and the soft massage ball is the smallest. The predictive accuracy is consistent with their stiffness. When the gripper is used to grasp a soft object with great deformation, the force value detected by the tactile sensor may become smaller. After converting the tactile readings to pictures, the image sequences for the stable state may be very similar to slip, which leads to misprediction.

\begin{figure*}[!t]
	\centering
	\includegraphics[width=15cm]{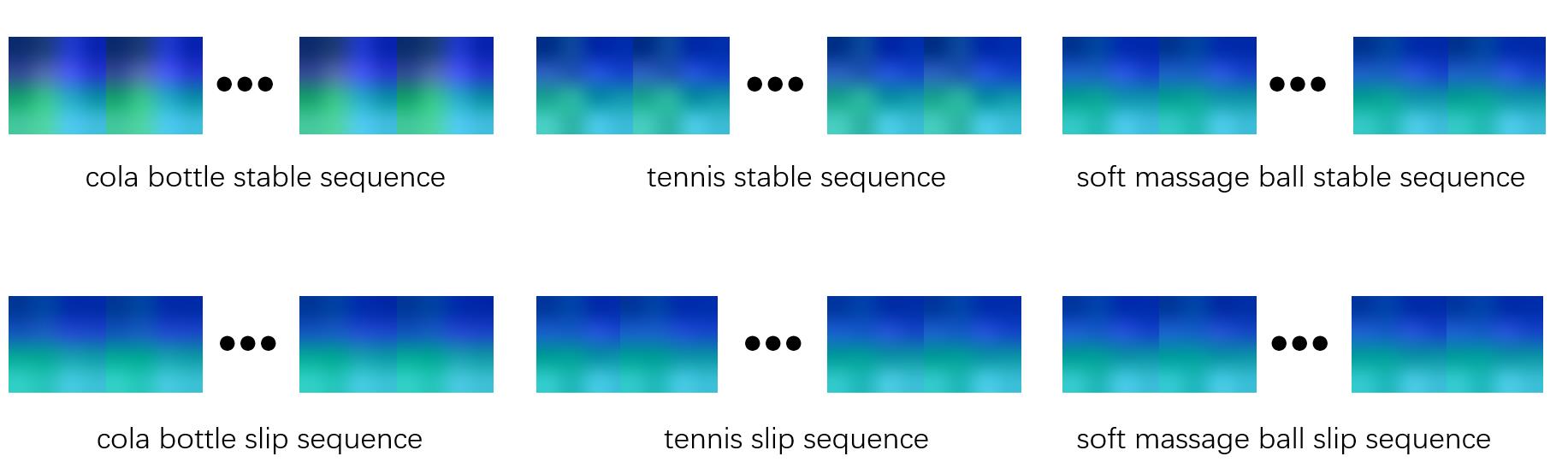} 
	\caption{The tactile image sequences: Stable grasp (top), sliding (bottom). cola bottle (left), tennis ball (centre), soft massage ball (right).}
	\label{fig:test tactile sequence}
\end{figure*}

To further verify this hypothesis, the stable and slip image sequences are randomly selected from the tactile image sequences of these three objects, which are shown in Fig. \ref{fig:test tactile sequence}. The stable tactile pictures are obviously different from the slips for the cola bottle. However, for the soft massage ball, the stable pictures are very similar to the slips. Although the stable pictures of the tennis ball are similar to the slips, the difference can be still seen. This may be the reason why the prediction accuracy varies greatly for objects with different stiffness.

\section{Conclusion}
\label{sec:conclusion}
Humans can recognize the grasping state through visual and tactile perception naturally. Inspired by this, taking the advantage of the visual and tactile information to help robots perform slip detection in grasping tasks, a novel visuo-tactile fusion model (CNN-MSTCN) is proposed to for slip detection when robots grasp objects. A total of 952 grasping and lifting experiments on 50 daily objects are performed using a 7-DoF robot manipulator equipped with XELA tactile sensors and a D455 RGB camera to collect tactile and visual data. %We have released our public dataset. 
The slip detection accuracy by using the proposed model can reach up to 96.96\%, where the grasping data of 40 objects are used for model training, and the other 10 for testing. The experiments on different input modalities demonstrate that the visuo-tactile fusion can outperform any other single modal input. By comparing with other fusion models, the proposed CNN-MSTCN can fuse the visual and tactile modal features and improve the detection performance more effectively. To verify the generalization ability of the proposed model, comparison experiments on the dataset collected by the GelSight tactile sensor have been also conducted. The results show that the proposed model has good generalization performance, i.e., it can be used for either array-shaped tactile sensors or optical tactile sensors. It has significant meaning in deploying the model to the scenarios using different tactile sensors. The proposed model can help robots perform slip detection, and enable automatic adjustment of grasping strategies. Meanwhile, it can provide a new idea for spatiotemporal feature extraction and modal fusion on visuo-tactile information.

In addition, it can be found that the tactile modal does not perform well on slip detection in comparison with the visual modal. Robot predicts stiff objects more accurately than soft ones by tactile sensing, which often mistakes stable grasping for slip. To explore the underlying reasons, some specific research and analysis are implemented. Although the visuo-tactile fusion can address these issues to a certain extent, it still affects the model performance. In future, one possible research direction is to further improve the comprehensive performance of the proposed model using different methods to convert the tactile data into images or extract the features of tactile data directly.

\bibliographystyle{IEEEtran}
\bibliography{references}

\begin{IEEEbiography}[{\includegraphics[width=1in,height=1.25in,clip,keepaspectratio]{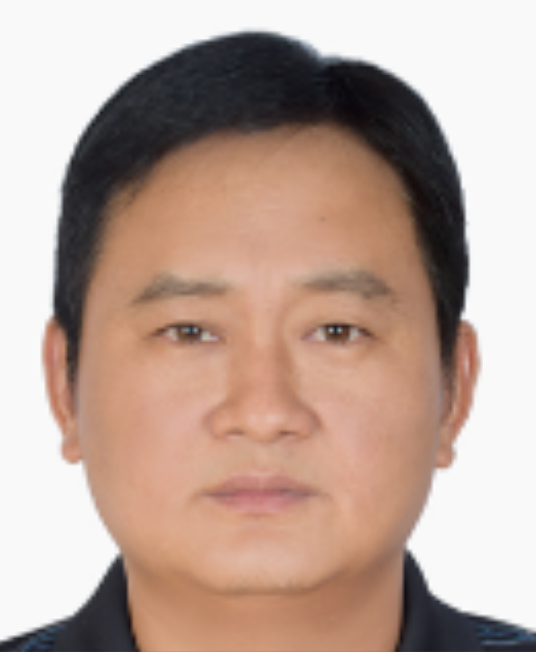}}]{Gao Junli}
received the M.Eng. degree in control theory and control engineering from Guangdong University of Technology, GuangZhou, China, in 2002; and the Ph.D. degree in mechanical manufacture and automation from South China University of Technology, GuangZhou, China, in 2005. 

He is an Associate Professor with Guangdong University of Technology, GuangZhou, China. His research interests include robot perceptin and manipulation control, motion control and embedded system application.
\end{IEEEbiography}

\begin{IEEEbiography}[{\includegraphics[width=1in,height=1.25in,clip,keepaspectratio]{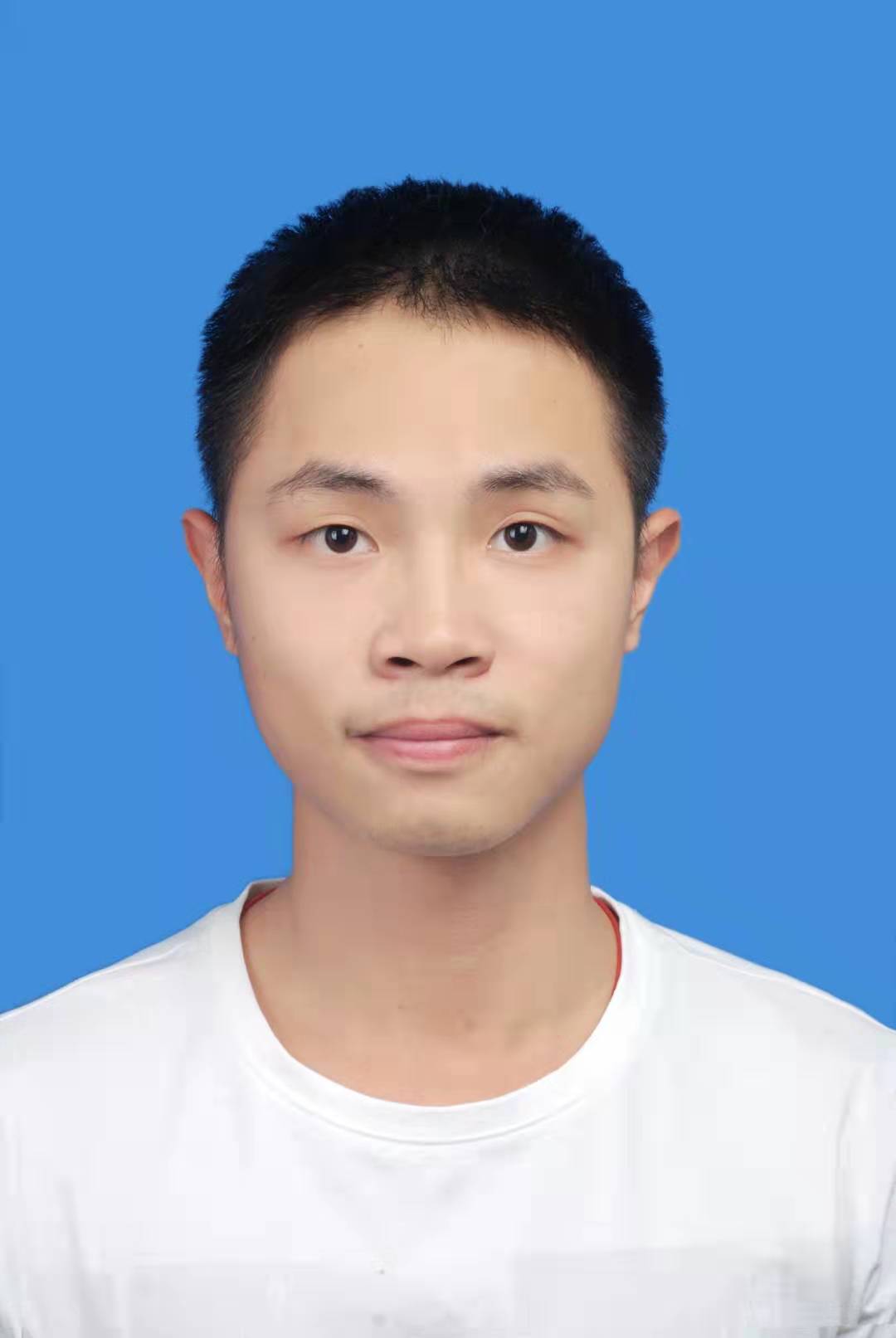}}]{Huang Zhaoji}
received the B.Eng. degree in electrical engineering and automation from Guangzhou University, GuangZhou, China, in 2016. He is currently working toward the M.A.Eng. degree with control science and engineering, Guangdong University of Technology, GuangZhou, China.

His research interests include robot grasping, robot control, and robot perception.
\end{IEEEbiography}

\begin{IEEEbiography}[{\includegraphics[width=1in,height=1.25in,clip,keepaspectratio]{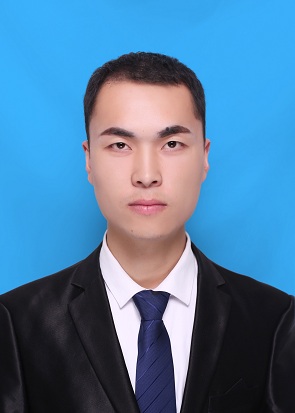}}]{Tang Zhaonian}
received the B.Eng. degree in measurement and control technology and instrument from Zhongyuan University of Technology, ZhengZhou, China, in 2015. He is currently working toward the M.Eng. degree with control engineering, Guangdong University of Technology, GuangZhou, China.

His research interests include robot control, robot grasping configuration, and machine vision.
\end{IEEEbiography}

\begin{IEEEbiography}[{\includegraphics[width=1in,height=1.25in,clip,keepaspectratio]{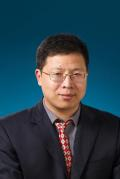}}]{Song Haitao}
received the B.S. degree in mathematics from Fuyang Normal College, Fuyang, China, in 1997; the M.Eng. degree in operations research and cybernetics from Nanjing Normal University, Nanjing, China, in 2004; and the Ph.D. degree in mechanical manufacture and automation from South China University of Technology, GuangZhou, China, in 2007. 

He is an Associate Professor with South China University of Technology, GuangZhou, China. His research interests include information management, data mining and intelligent decision-making.
\end{IEEEbiography}

\begin{IEEEbiography}[{\includegraphics[width=1in,height=1.25in,clip,keepaspectratio]{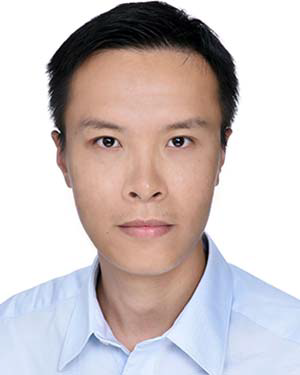}}]{Liang Wenyu}
(Member, IEEE) received the B.Eng. and M.Eng. degrees in mechanical engineering from the China Agricultural University, Beijing, China, in 2008 and 2010, respectively, and the Ph.D. degree in electrical and computer engineering from the National University of Singapore, Singapore, in 2014.

He is currently a Scientist with the Institute for Inforcomm Research, A*STAR, Singapore and also an Adjunct Assistant Professor with the Department of Electrical and Computer Engineering, National University of Singapore. His research interests mainly include robotics, intelligent systems, precision motion control, and force control.
\end{IEEEbiography}

\vfill

\end{document}